\documentclass[11pt]{article}
\usepackage{float} 
\usepackage{amssymb} 
\usepackage{amsmath} 
\usepackage{acl2016}
\usepackage{times}
\usepackage{url}
\usepackage{latexsym}
\usepackage{smartdiagram}
\usepackage{tikz}
\usetikzlibrary{calc}
\usepackage{qtree}  
\usepackage{multirow} 
\usepackage{xcolor,colortbl} 
\usepackage{array,etoolbox}
\usepackage{gb4e} 
\usepackage[colorinlistoftodos]{todonotes}

\makeatletter
\newcommand\footnoteref[1]{\protected@xdef\@thefnmark{\ref{#1}}\@footnotemark}
\makeatother
\preto\tabular{\setcounter{magicrownumbers}{0}}
\newcounter{magicrownumbers}

\def\itemrange#1{%
\addtocounter{enumi}{1}%
\edef\labelenumi{\theenumi--\noexpand\theenumi.}%
\addtocounter{enumi}{-1}%
\addtocounter{enumi}{#1}%
\item
\def\labelenumi{\theenumi}}

\aclfinalcopy 

\title{The CLaC Discourse Parser at CoNLL-2016}

\author{Majid Laali\thanks{\:\:Both authors contributed equally} \hspace{2cm} Andre Cianflone\textsuperscript{$\ast$} \hspace{2cm} Leila Kosseim\\
  Department of Computer Science and Software Engineering \\
  Concordia University, Montreal, Quebec, Canada \\
  {\tt \{Laali, Cianflone, Kosseim\}@encs.concordia.ca} \\
  }
  
\date{}

\begin{document}
\maketitle

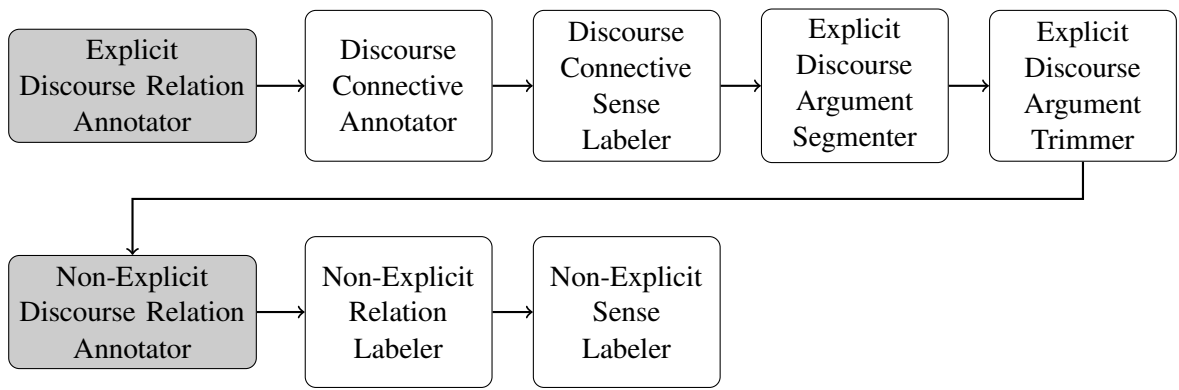
\begin{figure*}[!ht]
\centering
\begin{tikzpicture}

\tikzstyle{part} = [rectangle, rounded corners, minimum height=1.5cm, align=center, draw=black, fill=black!20, text width=3cm]
\tikzstyle{component} = [rectangle, rounded corners, minimum height=2cm, align=center, draw=black, text width=2.2cm]
\tikzstyle{arrow} = [thick, ->]

  \def \yspc {-2cm}
  \def \xpsc {2cm}
  \def \bigxpsc {2.5cm}
  
  \node[part] (P1) {Explicit \\ Discourse Relation Annotator};
  \node[component, right of=P1, xshift=\bigxpsc] (P11) {Discourse Connective Annotator};
  \node[component, right of=P11, xshift=\xpsc] (P12) {Discourse Connective Sense Labeler};
  \node[component, right of=P12, xshift=\xpsc] (P13) {Explicit Discourse Argument Segmenter};
  \node[component, right of=P13, xshift=\xpsc] (P14) {Explicit Discourse Argument Trimmer};
  \draw[arrow] (P1) -- (P11);
  \draw[arrow] (P11) -- (P12);
  \draw[arrow] (P12) -- (P13);
  \draw[arrow] (P13) -- (P14);

  \node[part, below of=P1, yshift=\yspc] (P2) {Non-Explicit Discourse Relation Annotator };
  \node[component, right of=P2, xshift=\bigxpsc] (P22) {Non-Explicit Relation Labeler};
  \node[component, right of=P22, xshift=\xpsc] (P23) {Non-Explicit Sense Labeler};
  \draw[arrow] (P14) |- ($(P1)!.5!(P2)$) -- (P2);
  \draw[arrow] (P2) -- (P22);
  \draw[arrow] (P22) -- (P23);
 
\end{tikzpicture}

\caption{Pipeline of the CLaC Discourse Parser}
\label{fig:clac:pipeline}
\end{figure*}

\begin{abstract}
 This paper describes our submission (\textit{CLaC}) to the CoNLL-2016 shared task on shallow discourse parsing. We used two complementary approaches for the task. A standard machine learning approach for the parsing of explicit relations, and a deep learning approach for non-explicit relations.
   Overall, our parser achieves an F\textsubscript{1}-score of 0.2106 on the identification of discourse relations (0.3110 for explicit relations and 0.1219 for non-explicit relations) on the blind CoNLL-2016 test set. 
\end{abstract}

\section{Introduction} \label{section:intro}

Shallow discourse parsing is defined as the identification of two discourse units, or discourse arguments, and labeling their relation. Although the topic of shallow discourse parsing has received much interest in the past few years (e.g. \cite{zhang15,weiss15,ji15,rutherford14,kong14,feng14}), the performance of the state-of-the-art discourse parsers is not yet adequate to be used in other downstream Natural Language Processing applications. For example, the best parser submitted at CoNLL-2015 \cite{wang15} achieved an F\textsubscript{1} score of 0.2400 on the blind test dataset.

\vspace{\baselineskip}
For the CoNLL 2016 task of shallow discourse parsing, four types of discourse relations have to be annotated in texts (more details of the task can be found in \cite{xue16}): 
\begin{enumerate}
\item \textit{Explicit Discourse Relations}: explicit discourse relations are explicitly signalled within the text through \textit{discourse connectives} such as \textit{because, however, since,} etc.
\item \textit{Implicit Discourse Relations}: implicit discourse relations are inferred by the reader and no discourse connective is used within the text to convey the relation. As a reader, implicit discourse relations can be inferred by inserting a discourse connective (called an \textit{implicit discourse connective}) in the text that best expresses the inferred relation.
\item \textit{AltLex Discourse Relations}: Similarly to implicit discourse relations, AltLex are not signalled through the presence of discourse connectives in the text. However, the relation is alternatively lexicalized by some non-connective expression, hence inserting an \textit{implicit discourse connective} to express the inferred relation would lead to a redundancy.
\item \textit{EntRel Discourse Relations}: EntRel discourse relations are defined between two discourse arguments where only an entity-based coherence relation could be perceived.
\end{enumerate}

In this paper, we report on the development and results of our discourse parser for the CoNLL 2016 shared task. As shown in Figure~\ref{fig:clac:pipeline}, our parser, named \textit{CLaC Discourse Parser}, consists of two main components: the Explicit Discourse Relation Annotator  and the Non-Explicit Discourse Relation Annotator . 

The Explicit Discourse Relation Annotator  is based on the parser that we submitted last year to CoNLL 2015 \cite{laali15}. For this year's submission, we improved its components by (1) adding new features (see Section~\ref{sec:features} for more details), (2) using a sequence classifier instead of a multiclass classifier in the Discourse Argument Segmenter, and (3) defining a new component, the Discourse Argument Trimmer, to identify  attributes and prune discourse arguments.

Last year's system did not address the annotation of non-explicit discourse relations (i.e. implicit, AltLex and EntRel discourse relations). For this year, we therefore built this module from scratch. The Non-Explicit Discourse Relation Annotator  first uses a binary Convolutional Neural Network (ConvNet) to detect whether a relation exists in a text devoid of a discourse connective, then uses a multiclass ConvNet to label the relation.


\section{Explicit Discourse Relation Annotator }
\label{sec:features}
\label{sec:explicit:pipeline}

Figure~\ref{fig:clac:pipeline} shows the pipeline of the CLaC parser. The top row in Figure~\ref{fig:clac:pipeline} focuses on the Explicit Discourse Relation Annotator. This pipeline consists of four main components: (1) \textit{Discourse Connective Annotator}, (2) \textit{Discourse Connective Sense Labeler}, (3) \textit{Explicit Relation Argument Segmenter} and (4) \textit{Discourse Argument Trimmer}.

Modules 1, 2 and 3 are based on last year's system \cite{laali15} while module 4 has been newly developed to address a weak issue from last year. 

\subsection{Discourse Connective Annotator}
\label{sec:connective:annotator}

The \textit{Discourse Connective Annotator} annotates discourse connectives within a text. To label discourse connectives, the annotator first searches the input texts for terms that  match any of the 100 discourse connectives listed in the Penn Discourse Treebank \cite{prasad08}. Inspired by \cite{pitler09-a}, a C4.5 decision tree binary classifier \cite{quinlan93} is used to detect if each discourse connective is used in a discourse usage or not. In addition to the six features proposed by \cite{pitler09-a}, this year we also used four of the features proposed by \cite{lin14}. In total 10 features were used:
\begin{enumerate}
	\item The discourse connective text in lowercase.
    \item The categorization of the case of the connective: \textit{all lowercase} or \textit{initial uppercase}.
    \item The highest node (called the \textit{SelfCat} node) in the parse tree that covers the connective words but nothing more.
    \itemrange{2} The parent, the left sibling and the right sibling of the \textit{SelfCat}.
	\itemrange{3} The left and the right word of discourse connective and their parts of speech.
\end{enumerate}

\subsection{Discourse Connective Sense Labeler} 

Once discourse connectives have been classified as discourse usage or not, the \textit{Discourse Connective Sense Labeler} labels the discourse relation signalled by the annotated discourse connectives with one of the 14 labels specified by the task. This component also uses a C4.5 decision tree classifier \cite{quinlan93} with the same 10 features used by the \textit{Discourse Connective Annotator} (see Section~\ref{sec:connective:annotator}).

\subsection{Discourse Argument Segmenter}
\label{sec:exp:argumnet:segmenter}
The goal of the \textit{Discourse Argument Segmenter} is to detect the discourse argument boundaries. This module first assumes that both discourse arguments (i.e. \textsc{Arg1} and \textsc{Arg2}) are located in the same sentence that contains the discourse connective. If \textsc{Arg1} is not found in the sentence, then the Discourse Argument Segmenter selects the immediately preceding sentence as \textsc{Arg1}. 


We used a similar approach proposed by \cite{kong14} to identify discourse arguments that appear in the same sentence. That is to say, we first select all the constituents in the parse tree that are directly connected to one of the nodes in the path from the discourse connective to the root of the sentence and classify them into to one of three categories: \textit{part-of-\textsc{Arg1}}, \textit{part-of-\textsc{Arg2}} or \textit{\textsc{Non}} (i.e. not part of any discourse argument). Then, all constituents which are tagged as part of \textsc{Arg1} or as part of \textsc{Arg2} are merged to obtain the actual boundaries of \textsc{Arg1} and \textsc{Arg2}. 

Instead of using integer programming as proposed by \newcite{kong14}, we used a Conditional Random Field (CRF) in order to leverage global information (i.e. information across all constituent candidates). CRFs have been previously used for discourse argument identification \cite{ghosh11} but at the token level. \newcite{kong14}'s approach generates a sequence of constituents and therefore, CRFs can be applied at the constituent level. 

We used the following categories of features for the CRF:
\begin{enumerate}
\item \textit{Discourse connective features:} This category includes all 10 features used in the Discourse Connective Annotator (see Section~\ref{sec:connective:annotator}).
\item \textit{Constituent features:} Motivated by \newcite{kong14}'features, we defined the following five features: 
\begin{enumerate}
\item The constituents in the path from the current constituent to the \textit{SelfCat} node in the parse tree.
\item The length of the path between the current constituent and the \textit{SelfCat} node.
\item The context of the current constituent in the parse tree. The context of a constituent is defined by its label, the label of its parent and the label of its left and right siblings in the parse tree.
\item The position of the current constituent relative to the \textit{SelfCat} node (i.e. left or right).
\item The syntactic production rule of the current constituent. 
\end{enumerate}
\item \textit{Lexical features:} This year, we also used lexical features including the head of the current constituent and four tokens that appear in the constituent boundary (the first token of the constituent and its previous token and the last token of the constituent and its following token). 
\end{enumerate}

\subsection{Discourse Argument Trimmer} 
According to the PDTB manual \cite{prasad08-a}, annotators should keep the span of two discourse arguments as small as possible and should remove any extra information that is not necessary for the discourse relation. Following this idea, the \textit{Discourse Argument Trimmer} is a classifier that excludes any constituent from the discourse argument span that is not related to the discourse relations. 

To do so, we developed a binary classifier that labels all the constituents and tokens in the annotated discourse arguments with either \textit{part-of-Argument} or \textit{Not-part-of-Argument} to exclude tokens that are not part of the discourse argument. Once the classifier has labeled all the tokens and constituents, we remove from the discourse arguments all tokens that are labeled as \textit{Not-Part-of-Argument} or part of a constituent with the \textit{Not-Part-of-Argument} label.

A C4.5 decision tree binary classifier was developed using the following features:
\begin{enumerate}
\item The head of the constituent or the text of the token.
\item The label of the constituent in the syntax tree or the POS of token.
\item The position of the constituent/token (i.e whether it appears at the beginning, inside or at the end of the discourse argument).
\item The syntactic production rule of the constituent's parent and grand parent or ``null'' for tokens.
\item The type of the argument (i.e. \textsc{Arg1} or \textsc{Arg2})
\item The node label/POS of the left and right siblings of the constituent/token in the syntactic tree. 
\end{enumerate}

\section{Non-Explicit Discourse Relation Annotator }
\label{sec:implicit:pipeline}
As mentioned in Section \ref{section:intro}, last year, the CLaC Discourse Parser did not address non-explicit relations. Therefore, for this year's participation we developped this module from scratch. Because Convolutional Neural Networks (ConvNets) have been successful at several sentence classification tasks (e.g. \cite{zhang2015sensitivity,kim:cnn}), we wanted to investigate if similar networks could be used to address the task of non-explicit discourse relation recognition.

The Non-Explicit Discourse Relation Annotator  begins where the Explicit Discourse Relation Annotator ends. The Explicit Discourse Relation Annotator only analyzes texts which contain a discourse connective; all other segments are sent to the Non-Explicit Discourse Relation Annotator. 

Because these text segments may or may not contain a discourse relation, the Non-Explicit Discourse Relation Annotator first sends each text segment to a binary ConvNet to identify which segments contain a discourse relations and which do not. The Non-Explicit Discourse Relation Annotator trims trailing discourse punctuation as per the shared task requirement. Only discourses with two consecutive arguments are considered as possible non-explicit discourses. Non-discourse segments are removed from the pipeline. Sense labelling is then performed on the remaining segments using a multiclass ConvNet.

\subsection{Input}
The two ConvNets have an identical setup. The input to the models are pretrained word embeddings from the Google News set, as trained with Word2Vec\footnote{https://code.google.com/archive/p/word2vec/}. Words not in the Google News set are randomly initialized. Word embeddings are non-static, meaning that they are allowed to change during training.

Each input to the networks is composed of the two padded discourse arguments. \textsc{Arg1} is padded to the length of the longest \textsc{Arg1}, and \textsc{Arg2} is similarly padded to the length of the longest \textsc{Arg2}. Since the training set contains a few unusually long arguments, we limited the argument size to the size of the 99.5\textsuperscript{th} percentile. This reduced the length of \textsc{Arg1} from 1000 to 60 words, and that of \textsc{Arg2} from 400 to 61 words. This dramatically decreased the model complexity with insignificant impact on performance. The two arguments are then concatenated to form a single input. Each word is then replaced with their embedded vector representation. 

Let $l$ be the length of a single input (the number of words in the discourse plus padding, 121). Let $d$ be the dimensionality of a word vector (300 for our pretrained embedding). Then the input to the networks, the matrix of discourse embedding, can be denoted $Q \in \mathbb{R}^{l \times d}$.

\subsection{Network}
The network configuration is largely based on \cite{kim:cnn}. We applied a narrow convolution over $Q$ with height $w$ (i.e. $w$ words) and width $d$ (the entire word vector) defined as region $h \in \mathbb{R}^{d \times w}$. We added a bias $b$ and applied a nonlinear function $f$ on the convolution to give us features $c_i$, where $i$ is the i\textsuperscript{th} word in the discourse input. This is shown in Formula \ref{eq:c}.

\begin{equation} \label{eq:c}
c_i = f(h \cdot Q_{i:i+w-1} + b)
\end{equation}

The nonlinear function $f$ in our case was the exponential linear unit (ELU) \cite{elu}, indicated in Formula \ref{eq:elu}.
\begin{equation} \label{eq:elu}
f(x) = \begin{cases}
	x & \text{if $x>0$}\\
    \alpha(exp(x)-1) & \text{if $x \leq 0$}
\end{cases}
\end{equation}

Since the convolution is narrow, there are $l-w+1$ such features, giving us a \textit{feature map} $c \in \mathbb{R}^{l-w+1}$. We applied max-over-time pooling on $c$ to extract the most ``important'' feature as in Formula \ref{eq:max}.
\begin{equation} \label{eq:max}
y = max(c)
\end{equation}

We applied 128 feature maps and pooled each one of these. We repeated the entire process 3 times for $w=3, 4$ and $5$, and concatenated them together. This gave us a final matrix $M \in \mathbb{R}^{3 \times 128}$. We reshaped $M$ to a flat vector and applied dropout as our regularization \cite{srivastava2014dropout}, giving us vector $u \in \mathbb{R}^{384}$. $u$ is fully connected to a softmax output layer where loss is measured with cross-entropy. The network was trained in mini-batches and optimized with the \textit{Adam} algorithm \cite{kingma2015method}.
\section{Results and Analysis}
\label{sec:experiments}

\definecolor{LightGray}{gray}{0.85}
\begin{table*}[ht!]
\centering
\begin{tabular}{|l|cc|cc|cc|}
\cline{2-7}
 \multicolumn{1}{c|}{}& \multicolumn{2}{c|}{Development Dataset} & \multicolumn{2}{c|}{Test Dataset}             & \multicolumn{2}{c|}{Blind Test Dataset}             \\ 
 \multicolumn{1}{c|}{}& \multicolumn{2}{c|}{(PDTB)} & \multicolumn{2}{c|}{(PDTB)}             & \multicolumn{2}{c|}{(Wikinews)}             \\ \cline{2-7}
 \multicolumn{1}{c|}{} & CLaC         & Best (2015) & CLaC         & Best (2015) & CLaC               & Best (2015) \\ \hline 
\rowcolor{LightGray}
\multicolumn{7}{|l|}{\textit{Full Parsing}} \\ \hline
Overall                             & \textbf{0.3260}      & 0.3851       & \textbf{0.2442}      & 0.2499        & \textbf{0.2106}      & 0.2400      \\
Explicit                            & 0.4457      & 0.4977       & 0.3572      & 0.3447        & 0.3110      & 0.3038      \\
Non-Explicit                            & 0.2167      & 0.2876       & 0.1395      & 0.1511        & 0.1219      & 0.1887      \\
\hline
\rowcolor{LightGray}
\multicolumn{7}{|l|}{\textit{Identification of Explicit Discourse Connective}} \\ \hline
Explicit                            & 0.9203      & 0.9514       &    0.9100   &    0.9421     & \textbf{0.9020}      & 0.9186      \\ \hline

\rowcolor{LightGray}
\multicolumn{7}{|l|}{\textit{Argument Identification}} \\ \hline
Overall                             & 0.4929      & 0.5704       & 0.4173      & 0.4377        & 0.3912      & 0.4637      \\
Explicit                            & 0.4867      & 0.5352       & 0.4023      & 0.3882        & \textbf{0.3989}      & 0.4135      \\
Non-Explicit                            & 0.4987      & 0.6014       & 0.4311      & 0.4881        & \textbf{0.3844}      & 0.5041      \\ \hline
\rowcolor{LightGray}
\multicolumn{7}{|l|}{\textit{Sense Labeling (Supplementary task)}}\\ \hline
Overall                             & 0.6222      &  \multicolumn{1}{c|}{-}       & 0.5736      & 0.6802        & 0.5000      & 0.6327      \\
Explicit                            & 0.9074      & \multicolumn{1}{c|}{-}       & 0.8948      & 0.9079        & \textbf{0.7622}      & 0.7685      \\
Non-Explicit                            & 0.3712      & \multicolumn{1}{c|}{-}      & 0.2813      & 0.4734        & \textbf{0.2772}      & 0.5176      \\
\hline
\end{tabular}
\caption{F\textsubscript{1}-score of the CLaC Discourse Parser and the best parser of 2015 with Different Datasets.}
\label{table:results}
\end{table*}

Table~\ref{table:results} shows the F\textsubscript{1} scores of the CLaC  Discourse Parser and the best parser at CoNLL 2015 \cite{wang15} for different datasets. The overall F\textsubscript{1} score of the CLaC  parser is 0.2106 with the blind test dataset which is lower than the F\textsubscript{1} score of the best parser at CoNLL 2015 (i.e. 0.2400). For explicit relations, the performance of our parser (F\textsubscript{1}=0.3110) is higher than the performance of last year's best parser (F\textsubscript{1}=0.3038); however, for non-explicit relations there is gap between the performance of our parser (F\textsubscript{1}=0.1219) and the performance of last year's best parser (F\textsubscript{1}=0.1887). 


\subsection{Explicit Discourse Relation Annotator }
Table~\ref{table:results} shows that the argument segmentation component is the bottleneck of the Explicit Discourse Relation Annotator. While the CLaC Discourse parser achieves competitive results in the identification of explicit discourse connectives (F\textsubscript{1}=0.9020) and labeling the sense signalled by the discourse connectives (F\textsubscript{1}=0.7622) with the blind test dataset, its performance is rather low (F\textsubscript{1}=0.3989) for the identification of the discourse argument boundaries. 

Our results show that the CLaC Discourse Parser has difficulty in detecting
\textsc{Arg1}. As Table~\ref{table:arg:results} shows, the precision and recall for the identification of \textsc{Arg1} (i.e. P=0.4928 and R=0.4749) are significantly lower than for \textsc{Arg2} (i.e. P=0.7194 and R=0.6932). \textsc{Arg2} is syntactically bound to discourse connectives and therefore, it is easier to detect its boundaries. Moreover, as mentioned in Section~\ref{sec:exp:argumnet:segmenter}, our approach does not account for arguments that appear in non-adjacent sentences. However, according to \newcite{prasad08}, 9.02\% of \textsc{Arg1} in the PDTB do not appear in the sentence adjacent to the discourse connective.

\begin{table}[h]
\centering
\begin{tabular}{|l|rrr|}
\cline{2-4}
\multicolumn{1}{c|}{}			 & \multicolumn{1}{c}{\textbf{P}}      & \multicolumn{1}{c}{\textbf{R}}      & \multicolumn{1}{c|}{\textbf{F\textsubscript{1}}} \\ \hline
\textbf{Arg1}         & 0.4928 & 0.4749 & 0.4837 \\
\textbf{Arg2}         & 0.7194 & 0.6932 & 0.7061 \\
\textbf{Arg1 \& Arg2} & 0.4065 & 0.3917 & 0.3989 \\ \hline
\end{tabular}

\caption{Results of the CLaC Discourse parser for the identification of discourse arguments with the blind test dataset (exact match).}
\label{table:arg:results}
\end{table}

The exact match of CoNLL is a strict evaluation measure for the argument identification. For example, in Sentence~(\ref{ex:strict:evaluation}), our parser did not detect the word `\textit{it}' (boxed) and therefore, accordingly to the exact match scoring schema, the boundaries of the discourse arguments are incorrect. 

\begin{exe}
	\ex \label{ex:strict:evaluation} \textit{The law does allow the RTC to borrow from the Treasury up to \$5 billion at any time}. \underline{Moreover}, \textbf{\framebox[1.5\width]{it} says the RTC's total obligations may not exceed \$50 billion, but that figure is derived after including notes and other debt, and subtracting from it the market value of the assets the RTC holds}.\footnote{This example is taken from the CoNLL development dataset.}
\end{exe}

Such cases where the CLaC parser misses the argument boundaries by only a few words (added or deleted) are frequent. For example, as Table~\ref{table:arg:partial:results} shows, if we evaluate the argument boundaries with the partial match metric defined in the CoNLL evaluator, the performance increases significantly. The partial match metric accepts the argument boundaries if 70\% of the tokens of the identified discourse arguments are correct. Using this metric, the F\textsubscript{1} score of the identification of \textsc{Arg1} and \textsc{Arg2} increases by 0.1917 and 0.0777 respectively.

\begin{table}[h]
\centering
\begin{tabular}{|l|rrr|}
\cline{2-4}
\multicolumn{1}{c|}{}			 & \multicolumn{1}{c}{\textbf{P}}      & \multicolumn{1}{c}{\textbf{R}}      & \multicolumn{1}{c|}{\textbf{F\textsubscript{1}}} \\ \hline
\textbf{Arg1}         & 0.6740 & 0.6768 & 0.6754 \\
\textbf{Arg2}         & 0.7695 & 0.7986 & 0.7838 \\
\textbf{Arg1 \& Arg2} & 0.6386 & 0.6667 & 0.6523 \\ \hline
\end{tabular}

\caption{Results of the CLaC Discourse parser for the identification of discourse argument with the blind test dataset (partial match).}
\label{table:arg:partial:results}
\end{table}

We also observed that the Explicit Discourse Argument Trimmer has a difficulty detecting what parts of the texts are related to discourse relations especially if multiple events appear in the text with a \textsc{Temporal} discourse relation. For example, in Sentence~(\ref{ex:removed:info}) the parser identified the boxed words as \textsc{Arg1} and missed required information. On the other hand in Sentence~(\ref{ex:extra:info}) the parser included extra information in \textsc{Arg1}. This type of error appears more frequently for \textsc{Arg1} which explains why the partial match metric improves the identification of \textsc{Arg1} more than the identification of \textsc{Arg2}.

\begin{exe}
	\ex \label{ex:removed:info} \textit{We would have to wait \framebox[1.1\width]{until we have} \framebox[1.1\width]{collected on those assets}} \underline{before} \textbf{we can move forward}.
\end{exe}

\begin{exe}
	\ex \label{ex:extra:info} \framebox[1.05\width]{But the RTC also requires ``working''} \framebox[1.05\width]{capital \textit{to maintain the bad assets of}} \framebox[1.1\width]{\textit{thrifts that are sold}}, \underline{until} \textbf{the assets can be sold separately}.
\end{exe}


\subsection{Non-Explicit Discourse Relation Annotator }

Table~\ref{table:results} shows that for the task of non-explicit sense labelling the Non-Explicit Discourse Relation Annotator achieves an F\textsubscript{1}-score of 0.2813 on the test dataset and 0.2772 on the blind dataset, versus 0.3712 on the developement dataset. The similar performance on the test and blind datasets and the 10\% difference with the development dataset suggest overfitting of our neural network.

For argument segmentation, just removing tailing punctuations from consecutive sentences achieves an F\textsubscript{1}-score of 0.3884. According to \newcite{prasad08}, non-explicit relations are present between successive pairs of sentences within paragraphs, but also intra-sententially between complete clauses separated by a semicolon or a colon. Our simple argument segmentation heuristic ignores intra-sentential arguments. We believe that this accounts for its poor performance on the identification of discourse arguments.

When looking more closely at the sense labelling performance (data not shown), it seems that our network tends to overweight a few high prior probability senses, notably \textit{EntRel} and \textit{Expansion.Conjunction}. \textit{EntRel} is predicted for 46\% of samples, whereas it only represents 29\% of the development dataset. \textit{Expansion.Conjunction} is predicted for 24\% of samples, whereas it represents only 17\% of the development dataset.

We believe that one of the key issues for the Non-Explicit Discourse Relation Annotator is the size of the training set for non-explicit discourse. 17,813 samples is limited for a ConvNet, hence reducing the possible complexity of our model. The Non-Explicit Discourse Relation Annotator underperformed the best parser from CoNLL-2015 on sense labeling by 24.04\% for the blind dataset, showing the advantage of non-neural network machine learning techniques when training data is scarce.

\section{Conclusion and Future Work}
\label{sec:conclusion}


A major area of concern in our system is the argument identification, both for explicit and non-explicit discourse relations. If we compare the results of the Supplementary task and Full Parsing task in Table \ref{table:results}, we can see that the Full Parsing F\textsubscript{1}-scores are about half of the Supplementary task F\textsubscript{1}-scores due to mis-identification of arguments.

It is necessary to consider cases where \textsc{Arg1} appears in non-adjacent sentences to improve the identification of discourse arguments for explicit relations. We believe that by considering co-references in texts, we can expand our approach to address non-adjacent discourse arguments. Furthermore, it would be interesting to define new features by using \textsc{Arg2} to detect what information can be added to or removed from \textsc{Arg1}. Finally, we believe that new ways to identify discourse arguments, such as Recurrent Neural Networks (Long Short Term Memory), could enhance the performance of the argument identification. To improve the identification of discourse arguments for non-explicit relations, we plan to expand the Explicit Discourse Argument Trimmer for non-explicit relations.

For non-explicit sense labeling, we would like to experiment with a larger training set possibly by automatically expanding it.

\section*{Acknowledgments}
The authors would like to thank Sohail Hooda for his programming help and the anonymous reviewers for their comments on the paper. This work was financially supported by an NSERC Discovery Grant.
\bibliographystyle{acl2016}
\bibliography{discourse-andre,discourse-majid,general}
\end{document}